\pdfoutput=1

\documentclass[11pt]{article}

\usepackage[final]{acl}

\usepackage{times}
\usepackage{latexsym}

\usepackage{graphicx}
\usepackage{multirow}
\usepackage{colortbl}
\usepackage{subcaption}
\usepackage{caption}
\usepackage{xspace}
\usepackage{makecell}
\newcommand{\seeda}{{\sc{SEEDA}}\xspace}
\newcommand{\mtwo}{$M^{2}$\xspace}
\newcommand{\ptmtwo}{PT-$M^{2}$\xspace}

\usepackage[T1]{fontenc}

\usepackage[utf8]{inputenc}

\usepackage{microtype}

\usepackage{inconsolata}

\title{Large Language Models Are State-of-the-Art Evaluator \\for Grammatical Error Correction}

\author{
  Masamune Kobayashi$^\diamond$
  \ \ \
  Masato Mita$^\dagger$$^\diamond$
  \ \ \
  Mamoru Komachi$^\ddagger$
  \\
  $^\diamond$Tokyo Metropolitan University, Japan
  \ \ \
  $^\dagger$CyberAgent Inc.
  \ \ \
  $^\ddagger$Hitotsubashi University, Japan
  \\
  \texttt{kobayashi-masamune@ed.tmu.ac.jp,}
  \ \ \
  \texttt{mita\_masato@cyberagent.co.jp,}
  \\
  \texttt{mamoru.komachi@r.hit-u.ac.jp}
}

\begin{document}
\maketitle
\begin{abstract}
Large Language Models (LLMs) have been reported to outperform existing automatic evaluation metrics in some tasks, such as text summarization and machine translation. 
However, there has been a lack of research on LLMs as evaluators in grammatical error correction (GEC). 
In this study, we investigate the performance of LLMs in English GEC evaluation by employing prompts designed to incorporate various evaluation criteria inspired by previous research.
Our extensive experimental results demonstrate that GPT-4 achieved Kendall's rank correlation of 0.662 with human evaluations, surpassing all existing methods.
Furthermore, in recent GEC evaluations, we have underscored the significance of the LLMs scale and particularly emphasized the importance of fluency among evaluation criteria.

\end{abstract}

\section{Introduction}
Large Language Models (LLMs) have surpassed existing systems in various NLP tasks, showcasing their high capabilities of language understanding and generation~\cite{ye:arxiv2023,bubeck:arxiv2023}.
These LLMs, which have had a significant impact on recent NLP research, also demonstrate the ability to produce high-quality corrections in grammatical error correction (GEC)~\cite{schick:arxiv2022,dwivediyu:arxiv2022,fang:arxiv2023,loem:bea2023,coyne:arxiv2023}.

In recent years, several studies have been conducted on the use of LLMs as an evaluator.
In text summarization, dialogue generation, and machine translation, GPT-4 has demonstrated superior performance compared to existing automatic evaluation metrics~\cite{liu:arxiv2023,kocmi:eamt2023}.
While there is very little research on GEC evaluation, considering GPT-4's ability to explain grammatical errors with 90\% accuracy in human evaluations~\cite{song:arxiv2023}, it holds potential for evaluating corrections.
\citet{sottana-etal-2023-evaluation} conducted meta-evaluation using a limited number of systems, but there has been no comprehensive analysis using dozens of systems like traditional approaches such as \citet{grundkiewicz-etal-2015-human} and \citet{kobayashi2024revisiting}.

\begin{figure}[t]
\centering
\includegraphics[width=7.5cm]{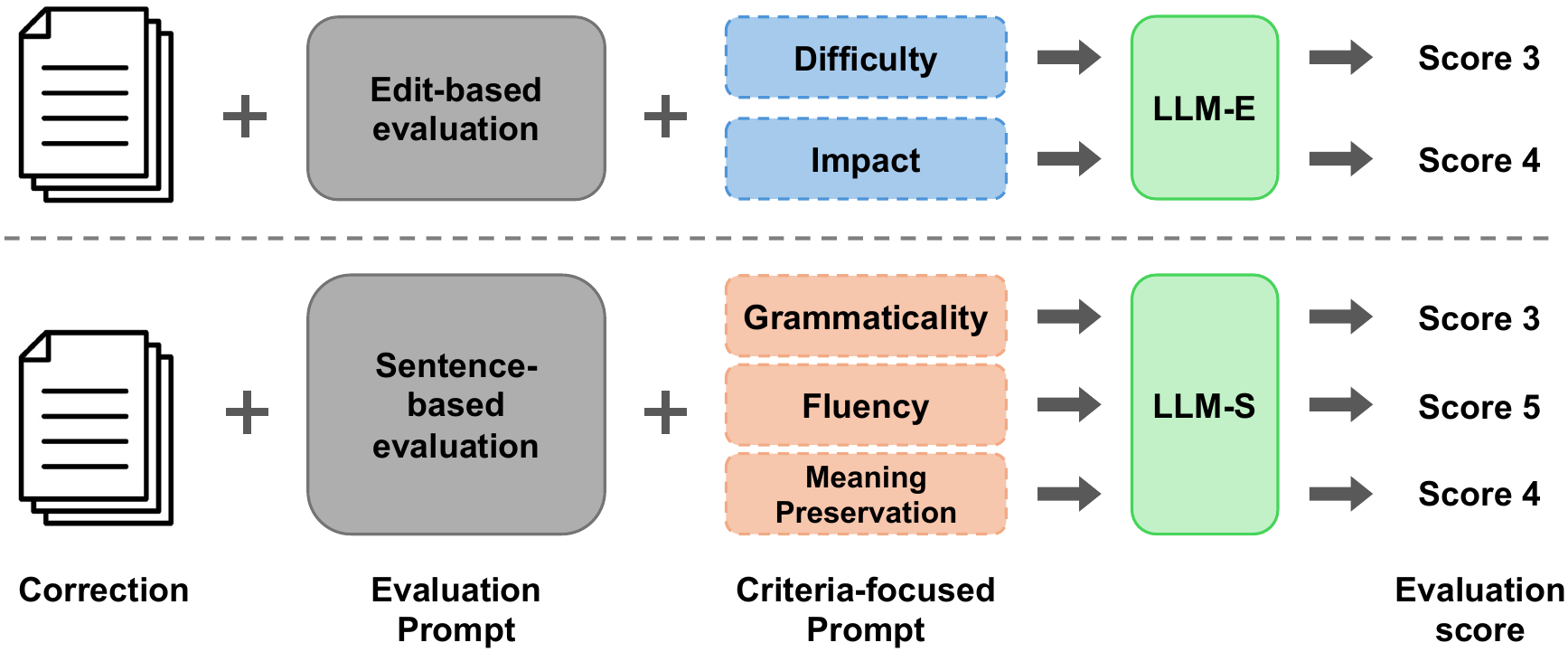} 
\caption{Evaluation framework using LLMs.}
\label{fig:punch}
\end{figure}

Therefore, we aim to explore the extent to which LLMs operate as evaluation models in English GEC.
Specifically, we conduct GEC evaluations using LLMs with prompts at different evaluation granularities to investigate how evaluation capabilities change with the presence of evaluation criteria and the scale of LLMs, as shown in Figure~\ref{fig:punch}.
\citet{kobayashi2024revisiting}'s work on the evaluation of metrics (i.e., meta-evaluation) has revealed that conventional metrics lack the resolution to capture performance differences in high-performing GEC systems.
Given this current state, to facilitate proper GEC evaluation moving forward, we investigate the potential of LLMs by comparing them with conventional metrics through meta-evaluation.

Our contributions are summarized as follows.
(1) We conducted a comprehensive investigation into the performance of LLMs as evaluators in GEC, and the results showed that GPT-4 achieved state-of-the-art performance, indicating the usefulness of considering evaluation criteria in prompts (especially fluency).
(2) It was suggested that as LLM scales decrease, the correlation with human evaluations decreases, and the ability to capture fluency in corrected sentences diminishes. Smaller LLMs tend to avoid extreme scores, while larger LLMs tend to assign higher scores.

\section{Experiment setup}
In this section, we explain the considered GEC metrics (\S \ref{section2.1}) and meta-evaluation methods (\S \ref{section2.2}).

\subsection{Considered metrics}\label{section2.1}
\paragraph{GEC metrics:}
We use two types of evaluation metrics: Edit-Based Metrics (EBMs), which assess only the edits made in the corrected text, and Sentence-Based Metric (SBMs), which evaluate the overall quality of the corrected sentences.

For EBMs, we employ four metrics.
\begin{itemize}
    \item \textbf{\mtwo~\cite{dahlmeier:acl2012}} dynamically extracts edits using Levenshtein algorithm to maximize overlap with gold annotations from the hypothesis sentences and calculates the F-score.
    \item \textbf{ERRANT~\cite{bryant:acl2017}} is similar to \mtwo, but it differs in that it uses a linguistically extended Damerau-Levenshtein algorithm for edit extraction to enhance the alignment of tokens with similar linguistic properties.
    \item \textbf{GoToScorer~\cite{gotou:coling2020}} calculates an F-score taking into account the difficulty of corrections.
    The difficulty is defined based on the number of systems that could correctly correct errors per total number of systems.
    \item \textbf{\ptmtwo~\cite{gong:emnlp2022}} combines \mtwo with BERTScore~\cite{DBLP:journals/corr/abs-1904-09675}, enabling the measurement of semantic similarity in addition to simply comparing edits.
\end{itemize}

For SBMs, we utilize four metrics.
\begin{itemize}
    \item \textbf{GLEU~\cite{napoles:ijcnlp2015}} rewards $n$-grams in the hypothesis sentence that match the reference but are not in the source sentence while penalizing $n$-grams in the source that do not match the reference.
    We use GLEU without tuning~\cite{napoles:arxiv2016}. 
    \item \textbf{Scribendi Score~\cite{islam:emnlp2021}} evaluates based on GPT-2 perplexity, token sort ratio, and Levenshtein distance ratio.
    \item \textbf{SOME~\cite{yoshimura:coling2020}} fine-tunes BERT~\cite{devlin:naacl2019} using human evaluation scores based on three criteria: grammaticality, fluency, and meaning preservation.
    \item \textbf{IMPARA~\cite{maeda:coling2022}} utilizes a quality estimation model and a similarity model based on BERT to consider the impact of edits.
\end{itemize}

\paragraph{LLMs:}
We consider three LLMs: LLaMa 2~\cite{touvron：arxiv2023} (13B for chat), GPT-3.5~\cite{NEURIPS2022_b1efde53} (\texttt{gpt-3.5-turbo-1106}), and GPT-4~\cite{openai2023gpt4} (\texttt{gpt-4-1106-preview}), conducting evaluations using prompts to assess both edits and sentences separately.
LLMs for edit-based evaluation are denoted with ``-E'' at the end, while ones for sentence-based evaluation have ``-S'' at the end.
Furthermore, we created prompts focusing on GEC evaluation criteria to investigate the impact of prompts on evaluation performance, comparing them with the base prompt.
For simplicity, this experiment uses only GPT-4 as the base LLM architecture.
GPT-4-E, which evaluates edits, focuses on the difficulty of corrections~\cite{gotou:coling2020} and the impact of edits~\cite{maeda:coling2022}.
GPT-4-S, which evaluates sentences, uses prompts focusing on grammaticality, fluency, and meaning preservation~\cite{asano:ijcnlp2017,yoshimura:coling2020}.
Detailed information on each prompt is provided in Appendix~\ref{sec:prompt}.

\begin{table*}[ht]
\centering
\resizebox{16cm}{!}{
\begin{tabular}{l|cc|cc|cc|cc|cc|cc|cc|cc}
\Xhline{3\arrayrulewidth}
\multirow{4}{*}{Metric} & \multicolumn{8}{c|}{System-level} & \multicolumn{8}{c}{Sentence-level} \\
& \multicolumn{4}{c|}{\seeda-E} & \multicolumn{4}{c|}{\seeda-S} & \multicolumn{4}{c|}{\seeda-E} & \multicolumn{4}{c}{\seeda-S} \\
& \multicolumn{2}{c|}{Base} & \multicolumn{2}{c|}{+ Fluent corr.} & \multicolumn{2}{c|}{Base} & \multicolumn{2}{c|}{+ Fluent corr.} & \multicolumn{2}{c|}{Base} & \multicolumn{2}{c|}{+ Fluent corr.} & \multicolumn{2}{c|}{Base} & \multicolumn{2}{c}{+ Fluent corr.} \\
& $r$ & $\rho$ & $r$ & $\rho$ & $r$ & $\rho$ & $r$ & $\rho$ & Acc & $\tau$ & Acc & $\tau$ & Acc & $\tau$ & Acc & $\tau$ \\
\Xhline{3\arrayrulewidth}
 \mtwo & 0.791 & 0.764 & -0.239 & 0.161 & 0.658 & 0.487 & -0.336 & -0.013 & 0.582 & 0.328 & 0.527 & 0.216 & 0.512 & 0.200 & 0.496 & 0.170 \\
 ERRANT & 0.697 & 0.671 & -0.502 & 0.051 & 0.557 & 0.406 & -0.587 & -0.116 & 0.573 & 0.310 & 0.511 & 0.188 & 0.498 & 0.189 & 0.471 & 0.129 \\
 GoToScorer & 0.901 & 0.937 & 0.667 & 0.916 & 0.929 & 0.881 & 0.627 & 0.881 & 0.521 & 0.042 & 0.505 & 0.009 & 0.477 & -0.046 & 0.504 & 0.009 \\
 \ptmtwo & 0.896 & 0.909 & -0.083 & 0.442 & 0.845 & 0.769 & -0.162 & 0.336 & 0.587 & 0.293 & 0.542 & 0.200 & 0.527 & 0.204 & 0.528 & 0.180 \\
\hline
 GLEU & 0.911 & 0.897 & 0.053 & 0.482 & 0.847 & 0.886 & -0.039 & 0.475 & 0.695 & 0.404 & 0.630 & 0.266 & 0.673 & 0.351 & 0.611 & 0.227 \\
 Scribendi Score & 0.830 & 0.848 & 0.721 & 0.847 & 0.631 & 0.641 & 0.611 & 0.717 & 0.377 & -0.196 & 0.359 & -0.240 & 0.354 & -0.238 & 0.345 & -0.264 \\
 SOME & 0.901 & 0.951 & 0.943 & 0.969 & 0.892 & 0.867 & 0.931 & 0.916 & 0.747 & 0.512 & 0.743 & 0.494 & 0.768 & 0.555 & 0.760 & 0.531 \\
 IMPARA & 0.889 & 0.944 & 0.935 & 0.965 & 0.911 & 0.874 & 0.932 & 0.921 & 0.742 & 0.502 & 0.725 & 0.455 & 0.761 & 0.540 & 0.742 & 0.496 \\
\Xhline{3\arrayrulewidth}
 GPT-3.5-E & -0.059 & 0.182 & -0.844 & -0.257 & -0.270 & -0.245 & -0.900 & -0.525 & 0.463 & -0.073 & 0.428 & -0.143 & 0.487 & -0.026 & 0.437 & -0.126 \\
 GPT-4-E & 0.911 & 0.965 & 0.845 & 0.974 & 0.839 & 0.846 & 0.786 & 0.899 & 0.728 & 0.455 & 0.702 & 0.404 & 0.698 & 0.395 & 0.687 & 0.374 \\
 \;+ Difficulty & 0.941 & 0.972 & 0.909 & 0.978 & 0.885 & 0.860 & 0.863 & 0.908 & 0.719 & 0.437 & 0.708 & 0.417 & 0.717 & 0.434 & 0.703 & 0.406 \\
 \;+ Impact & 0.905 & \textbf{0.986} & 0.848 & \textbf{0.987} & 0.844 & 0.860 & 0.793 & 0.908 & 0.730 & 0.460 & 0.710 & 0.420 & 0.717 & 0.434 & 0.696 & 0.392 \\
\hline
 Llama 2-S & 0.534 & 0.427 & 0.161 & 0.349 & 0.482 & 0.273 & 0.090 & 0.235 & 0.521 & 0.042 & 0.527 & 0.054 & 0.534 & 0.068 & 0.526 & 0.052 \\
 GPT-3.5-S & 0.878 & 0.916 & 0.302 & 0.648 & 0.770 & 0.636 & 0.199 & 0.433 & 0.633 & 0.265 & 0.597 & 0.195 & 0.631 & 0.263 & 0.608 & 0.216 \\
 GPT-4-S & 0.960 & 0.958 & 0.967 & 0.969 & 0.887 & 0.860 & 0.931 & 0.908 & 0.798 & 0.595 & 0.783 & 0.565 & 0.784 & 0.567 & 0.770 & 0.540 \\
 \;+ Grammaticality & 0.961 & 0.937 & 0.981 & 0.956 & 0.888 & 0.867 & \textbf{0.953} & 0.912 & 0.807 & 0.615 & 0.804 & 0.607 & 0.796 & 0.592 & 0.788 & 0.577 \\
 \;+ Fluency & \textbf{0.974} & 0.979 & \textbf{0.981} & 0.982 & 0.913 & 0.874 & 0.952 & 0.916 & \textbf{0.831} & \textbf{0.662} & \textbf{0.812} & \textbf{0.624} & \textbf{0.819} & \textbf{0.637} & \textbf{0.797} & \textbf{0.594} \\
 \;+ Meaning Preservation & 0.911 & 0.960 & 0.976 & 0.974 & \textbf{0.958} & \textbf{0.881} & 0.952 & \textbf{0.925} & 0.813 & 0.626 & 0.793 & 0.587 & 0.810 & 0.620 & 0.792 & 0.584 \\
\Xhline{3\arrayrulewidth}
\end{tabular}}
\caption{
Results of system-level and sentence-level meta-evaluations.
GPT-4-S demonstrated higher performance compared to existing GEC metrics, showing the most improvement in correlation when focusing on fluency.
}
\label{tab:meta}
\end{table*}

\subsection{Meta-evaluation methods}\label{section2.2}
We conduct system-level and sentence-level meta-evaluations using \seeda dataset~\cite{kobayashi2024revisiting}.
\seeda consists of human evaluations at two different granularities: edit-based and sentence-based, for 12 outputs from neural-based GEC systems and 3 human-authored sentences.
The dataset comprises two components: \seeda-E based on edit-based evaluation and \seeda-S based on sentence-based evaluation.
In SEEDA, for correction pairs (A, B) sampled from these corrected sentence collections, three annotators provide 5-point scores for each granularity, resulting in 5347 pairwise judgments (A$>$B, A$=$B, A$<$B). 
Subsequently, human rankings (from 1st to 15th place) of systems are obtained from pairwise judgments using rating algorithms such as Trueskill~\cite{sakaguchi:acl2014} and Expected Wins~\cite{bojar-etal-2013-findings}.
We conduct two variations of meta-evaluation: ``Base'', which uses the 12 systems excluding outliers, and ``+ Fluent corr.'', which adds two fluent corrected sentences\footnote{In GEC, there are two types of edits: minimal edits, which make the minimum necessary corrections, and fluency edits, which aim to make the sentence more fluent.} additionally.

\paragraph{System-level meta-evaluation:}
In the system-level meta-evaluation, we utilize the system scores derived from human rankings of systems using TrueSkill~\cite{sakaguchi:acl2014}.
For metrics like SOME, where system-level scores cannot be directly calculated, we use the average of sentence-level scores as a substitute.
Additionally, for LLMs, we employ system scores derived from LLMs rankings (Appendix \ref{sec:llm_ranking}) similar to human rankings.
To measure the correlation between human evaluations and metric scores, we use Pearson correlation ($r$) and Spearman rank correlation ($\rho$).
To ensure proper correlation calculation, we use the set of sentences that humans evaluated to compute the metric scores.

\paragraph{Sentence-level meta-evaluation:}
In the sentence-level meta-evaluation, we use pairwise judgments from \seeda.
To investigate the proximity between human evaluations and metric scores, we employ Accuracy (Acc) and Kendall's rank correlation ($\tau$). 
Kendall ($\tau$) is valuable for assessing performance in common use cases where corrections are compared to each other.

\begin{figure*}[t]
\centering
\subfloat[\seeda-E]{\includegraphics[clip, width=16cm]{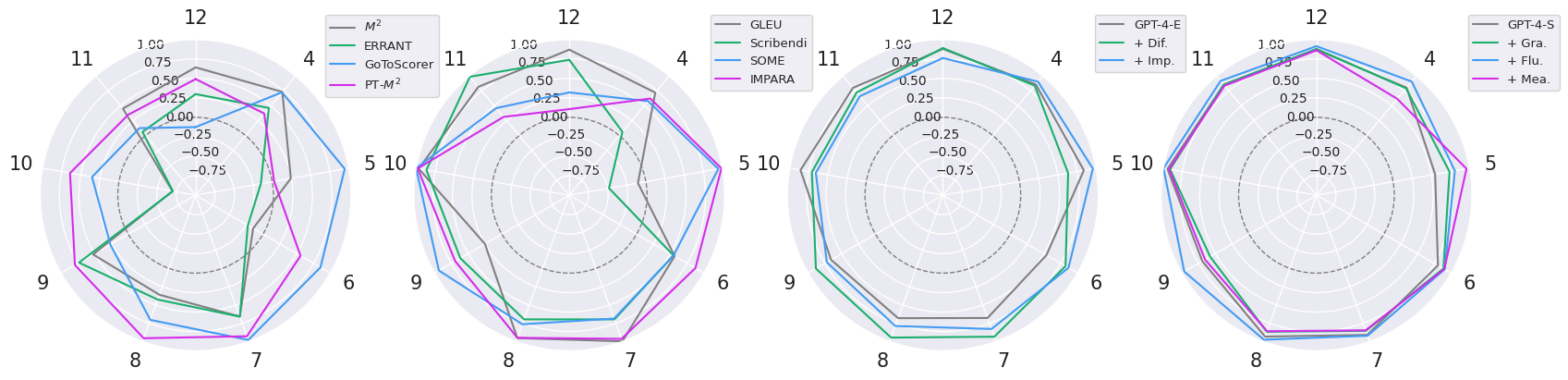}
\label{fig:rader_seedaE}}\\
\subfloat[\seeda-S]{\includegraphics[clip, width=16cm]{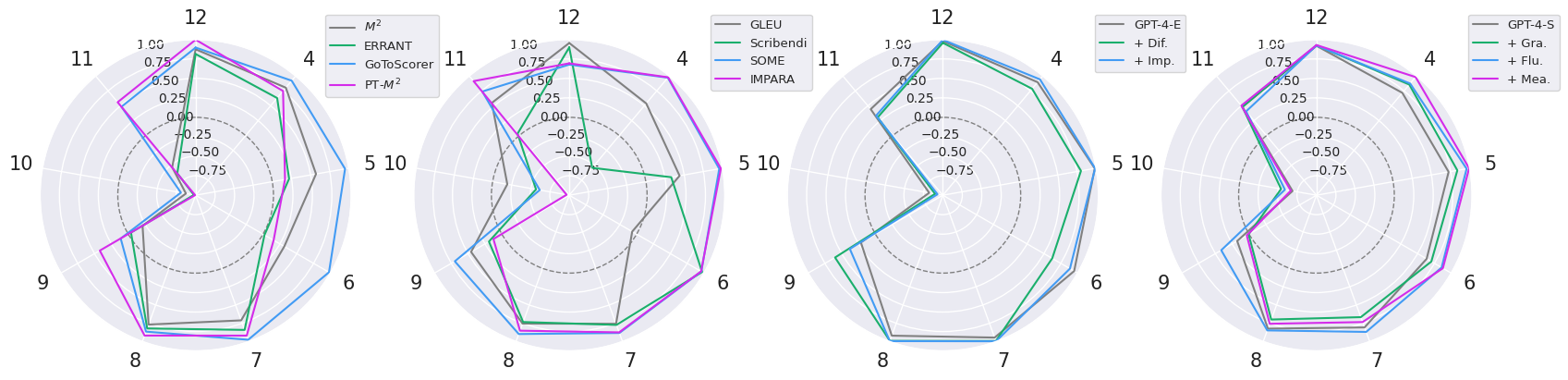}
\label{fig:radar_seedaS}}
\caption{
Window analysis was performed by selecting any consecutive four systems from the human rankings of the 12 systems (``Base'').
For instance, x=4 involves calculating the Pearson correlation ($r$) using the systems ranked from 1st to 4th in the human rankings.
In contrast to conventional GEC metrics, which exhibit unstable correlations, GPT-4 demonstrates relatively stable correlations.
}
\label{fig:radarchart}
\end{figure*}

\section{Results}
In this section, we analyze the performance of LLMs as GEC evaluators in system-level (\S \ref{section3.1}) and sentence-level meta-evaluations (\S \ref{section3.2}).
Additionally, we conduct further analysis by changing the system set to investigate the impact of the considered systems in the meta-evaluation(\S \ref{section3.3}).

\subsection{System-level analysis}\label{section3.1}
In Table~\ref{tab:meta},\footnote{Llama 2-E was excluded from this experiment because its output scores were not stable.} GPT-4 tends to achieve high correlations compared to existing metrics, highlighting their utility in GEC evaluations.
These prompts that focus on criteria tend to enhance correlation compared to base prompts, implying that GPT-4 can derive valuable insights from evaluation criteria. 
This observation aligns with recent studies that report performance improvements by incorporating additional sentences into the prompt~\cite{barham:arxiv2022,kojima:arxiv2023,li:arxiv2023}.

The decrease in correlation as the LLM scale decreases, such as with Llama 2 and GPT-3.5, suggests the importance of the LLM scale.
Especially, the decrease in correlation when adding fluent corrected sentences (``+ Fluent corr.'') compared to ``Base'' implies that smaller-scale LLMs may not adequately consider the fluency of sentences.
Possible reasons for this include issues such as LLM's tendency to produce the same scores (Appendix \ref{sec:llm_scoring}) and the inability to interpret the context of prompts as expected by users.
However, GPT-4 consistently demonstrated a high correlation and provided more stable evaluations compared to traditional metrics.

The fact that most system-level correlations for GPT-4 exceed 0.9 suggests that the conventional meta-evaluation using a dozen systems may have reached a performance saturation point for the task.
This poses a significant concern as it could lead to an underestimation of high-performing metrics in future meta-evaluations.
One possible solution is to utilize sentence-level correlations with a larger sample size or explore correlations between systems with similar performance levels, increasing the difficulty of the task.

\subsection{Sentence-level analysis}\label{section3.2}
In the sentence-level meta-evaluation, we observed differences in correlations between metrics that were not apparent in the system-level meta-evaluation.
In particular, while GPT-4-E and GPT-4-S showed similar correlations in system-level meta-evaluation, it was revealed that there was a notable difference between them.
Additionally, considering fluent corrected sentences (``+ Fluent corr.'') led to a slight decrease in overall correlation, but GPT-4 still maintained a considerably high correlation compared to traditional metrics.
This suggests that GPT-4 exhibits strong correlations with human evaluations and that examining sentence-level correlations is beneficial for comparing high-performance metrics.

Most prompts focused on criteria significantly improved sentence-level correlations compared to the base prompt.
Notably, GPT-4-S + Fluency demonstrated the ability to greatly enhance performance, surpassing existing GEC metrics and achieving state-of-the-art performance.
This suggests the need for a detailed examination of fluency beyond grammaticality when evaluating high-quality corrections. 
Paradoxically, it implies that humans also prioritize fluency when comparing high-quality corrected sentences.
Furthermore, the moderate fluctuations in correlation resulting from changing a single word in the prompt (GPT-4-S + Grammaticality vs. GPT-4-S + Fluency) highlight the impact of prompt engineering on performance.
In other aspects, the results were generally consistent with those in the system-level meta-evaluation.

\subsection{Further analysis}\label{section3.3}
To increase the difficulty of the meta-evaluation task, we computed correlations using a set of systems with similar performance.
Specifically, we conduct system-level meta-evaluation using only subsets of consecutive four systems in the human rankings of systems, and show the transitions of correlation at positions from 4th to 12th as window analysis in Figure~\ref{fig:radarchart}\footnote{For simplicity, we exclude the results of Llama 2 and GPT-3.5, which showed low performance.}.
For example, the point at x=4 represents the Pearson correlation value calculated using only the outputs of the four systems ranked from 1st to 4th.

According to the window analysis, GPT-4 maintains relatively high and stable correlations, making them suitable for evaluating modern neural systems in recent years.
In \seeda-E, the notably high correlations of GPT-4-S + Fluency across almost all data points emphasize the importance of fluency.
In \seeda-S, while overall correlations are high, the significant decrease in correlation at x=10, suggests the presence of GEC systems that are challenging to evaluate for the metrics.
On the other hand, conventional metrics frequently exhibit either no correlation or negative correlation, indicating their low robustness in GEC evaluation.

\section{Related Work}
Several studies have investigated the evaluation performance of LLMs.
\citet{chiang:acl2023} conducted the first investigation into LLM evaluation performance, demonstrating that GPT-3.5 can achieve expert-level evaluation in tasks such as open-ended story generation and adversarial attacks.
In the summarization task, \citet{liu:emnlp2023} revealed that GPT-4 has state-of-the-art evaluation performance by leveraging their proposed methods like auto-CoT (Chain-of-Thought) and weighted scores. 
In the machine translation task, \citet{kocmi:eamt2023} demonstrated that only larger models exceeding GPT-3.5 can perform translation quality evaluation, with GPT-4 slightly inferior to existing metrics at the segment level.
\citet{yancey:bea2023} utilized LLMs to evaluate second language writing proficiency through essay grading, discovering that GPT-4 exhibits performance equivalent to modern automated writing evaluation methods.


\section{Conclusion}
In this work, we investigated the capability of LLMs as evaluators in English GEC, and GPT-4 demonstrated significantly higher correlations compared to traditional metrics.
Future work should delve into the impact of few-shot learning and optimize prompt engineering for enhanced evaluation performance.
Furthermore, we plan to explore the possibility of document-level evaluation, considering the expansion of the GPT's context window, which is not currently focused on by existing metrics.


\section{Limitations}
Some of the LLMs (such as GPT-4) used in this study are not freely available and may require special access or payment to use. 
This could limit the applicability of our evaluation method. 
Additionally, since many LLMs are constantly updated, there is a possibility of inconsistent evaluation results across different versions.
To address this issue, we also conducted evaluations using reproducible LLMs (such as Llama 2).

\section*{Acknowledgments}
This work is partly supported by JST, PRESTO Grant Number JPMJPR2366, Japan.

\bibliography{custom}

\clearpage
\appendix
\section{Prompts for GEC evaluation}\label{sec:prompt}
\begin{figure*}[t]
\centering
\subfloat[Edit-based evaluation]{\includegraphics[width=8cm]{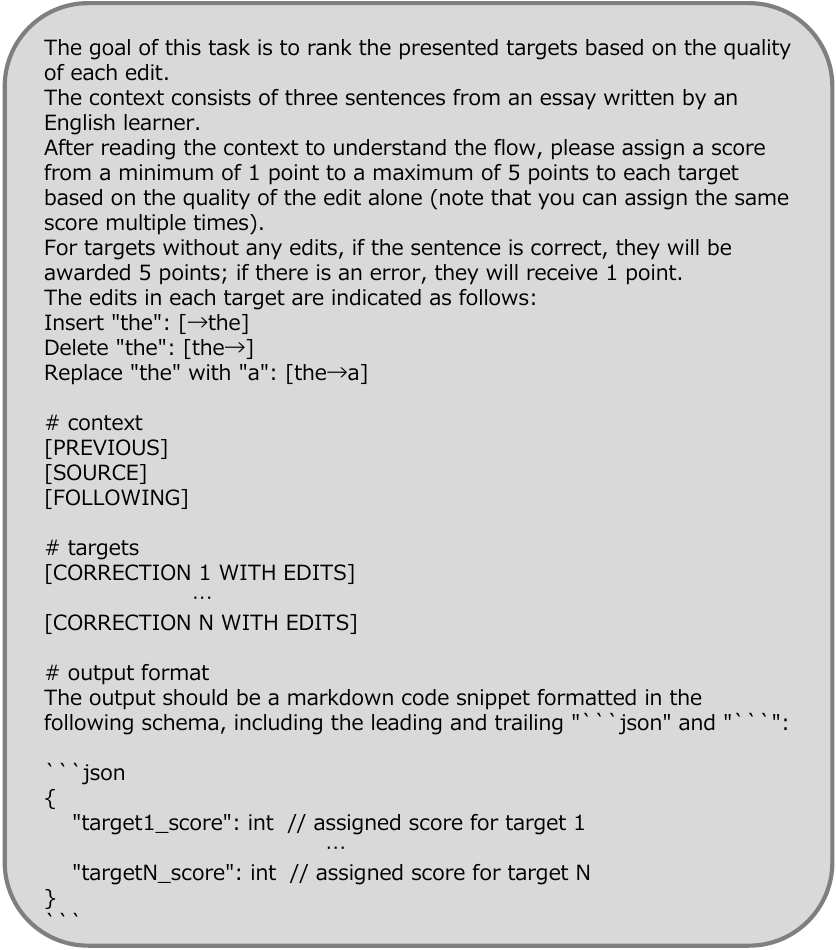}\label{fig:prompt_edit}}
\subfloat[Sentence-based evaluation]{\includegraphics[width=8cm]{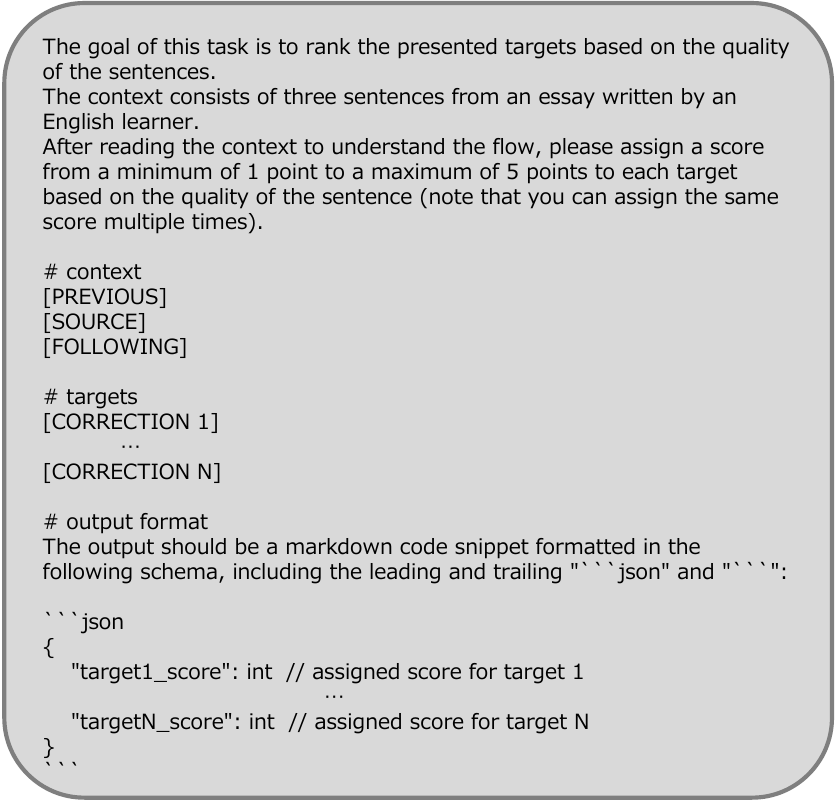}\label{fig:prompt_sent}}
\caption{Prompts used for edit-based evaluation and sentence-based evaluation by LLMs}
\label{fig:prompt}
\end{figure*}

The prompts used for edit-based evaluation and sentence-based evaluation by LLMs are illustrated in Figures~\ref{fig:prompt_edit} and ~\ref{fig:prompt_sent}, respectively.
In the \# context, [SOURCE] represents the source, and [PREVIOUS] and [FOLLOWING] are the preceding and succeeding sentences in the essay, respectively. 
In the \# targets, [CORRECTION N WITH EDITS] denotes a corrected sentence with explicitly indicated edits, while [CORRECTION N] represents a regular corrected sentence.
Here, N takes values from 1 to 5.
Additionally, the prompts output scores in JSON format to maintain a consistent output format.
For prompts focused on evaluation criteria, the following sentence is added to the end of the first paragraph of the prompt.
\begin{itemize}
    \item Difficulty: ``Please evaluate each edit in the target with a focus on the difficulty of corrections.''
    \item Impact: ``Please evaluate each edit in the target with a focus on its impact on the sentence.''
    \item Grammaticality: ``Please evaluate each target with a focus on the grammaticality of the sentence.''
    \item Fluency: ``Please evaluate each target with a focus on the fluency of the sentence.''
    \item Meaning Preservation: ``Please evaluate each target with a focus on preserving the meaning between each target and the source, which is the middle sentence in the context.''
\end{itemize}

An example of a prompt for evaluation using GPT-4-S + Fluency is provided below:
\\
\\
\textit{The goal of this task is to rank the presented targets based on the quality of the sentences.}
\\
\textit{The context consists of three sentences from an essay written by an English learner.}
\\
\textit{After reading the context to understand the flow, please assign a score from a minimum of 1 point to a maximum of 5 points to each target based on the quality of the sentence (note that you can assign the same score multiple times).}
\\
\textit{Please evaluate each target with a focus on the fluency of the sentence.}
\\
\\
\textit{\# context}
\\
\textit{These are the advantages that save works most of the time .}
\\
\textit{In conclude , socia media benefits people in several ways but in the same time harms people .}
\\
\textit{People should avoid the misuse of socia media and use it in the proper way .}
\\
\\
\textit{\# targets}
\\
\textit{In conclude , socia media benefits people in several ways but in the same time harms people .}
\\
\textit{In conclusion , social media benefits people in several ways but at the same time harms people .}
\\
\textit{In conclusion , social media benefits people in several ways but , at the same time , harms people .}
\\
\textit{In conclude , social media benefits people in several ways but at the same time harms people .}
\\
\textit{In conclusion , socia media benefits people in several ways but , at the same time , harms people .}
\\
\\
\textit{\# output format ...}
\\

\section{LLM rankings of GEC systems}\label{sec:llm_ranking}
The LLM rankings based on pairwise judgments (A$>$B, A$=$B, A$<$B) of corrections (A, B) conducted by LLMs and generated using Trueskill are shown in Table~\ref{tab:llm_rankings}.
It can be observed that LLMs with relatively smaller scales, such as GPT-3.5 and Llama2, have difficulty in ranking fluent corrections (REF-F and GPT-3.5) higher.
Furthermore, these LLMs tend to assign similar scores to many systems, suggesting that they may not effectively differentiate between the quality of corrections.
In contrast, GPT-4 can rank fluent corrections highly, resulting in rankings that closely resemble human evaluations.

\begin{table*}[ht]
  \centering
  \scalebox{0.7}[0.7]{
  \begin{minipage}[b]{0.4\textwidth}
    \centering
    \begin{tabular}{crcl}
    \Xhline{3\arrayrulewidth}
    \# & Score & Range & System \\
    \hline
    1 & 0.481 & 1     & INPUT \\
    \hline
    2 & 0.287 & 2     & UEDIN-MS \\
    \hline
    3 & 0.215 & 3-3   & GECToR-ens \\
    \hline
    4 & 0.110 & 4-6   & Riken-Tohoku \\
      & 0.089 & 4-8   & GECToR-BERT \\
      & 0.078 & 4-8  & TransGEC \\
      & 0.066 & 4-9  & PIE \\
      & 0.032 & 6-12 & REF-M \\
      & 0.025 & 7-12 & BERT-fuse \\
      & 0.017 & 7-13 & LM-Critic \\
      & -0.005 & 8-13 & BART \\
      & -0.008 & 8-13 & T5 \\
      & -0.011 & 9-13 & TemplateGEC \\
    \hline
    5 & -0.460 & 14   & GPT-3.5 \\
    \hline
    6 & -0.916 & 15   & REF-F \\
    \Xhline{3\arrayrulewidth}
    \end{tabular}
    \subcaption{GPT-3.5-E}
    \hspace{1mm}
  \end{minipage}
  }
  \hspace{7mm}
  \scalebox{0.7}[0.7]{
  \begin{minipage}[b]{0.4\textwidth}
    \centering
    \begin{tabular}{crcl}
    \Xhline{3\arrayrulewidth}
    \# & Score & Range & System \\
    \hline
    1 & 0.409 & 1     & GPT-3.5 \\
    \hline
    2 & 0.210 & 2-4   & REF-F \\
      & 0.182 & 2-4   & TransGEC \\
      & 0.148 & 3-6   & T5 \\
      & 0.127 & 3-7   & REF-M \\
      & 0.105 & 4-8   & BERT-fuse \\
      & 0.075 & 6-9  & UEDIN-MS \\
      & 0.071 & 6-9  & Riken-Tohoku \\
      & 0.064 & 6-9 & GECToR-BERT \\
      & 0.003 & 9-11 & PIE \\
      & -0.06 & 10-11 & LM-Critic \\
    \hline
    3 & -0.147 & 12-13 & TemplateGEC \\
      & -0.150 & 12-13   & GECToR-ens \\
    \hline
    4 & -0.266 & 14     & BART \\
    \hline
    5 & -0.770 & 15   & INPUT \\
    \Xhline{3\arrayrulewidth}
    \end{tabular}
    \subcaption{GPT-4-E}
    \hspace{1mm}
  \end{minipage}
  }
  \hspace{7mm}
  \scalebox{0.7}[0.7]{
  \begin{minipage}[b]{0.4\textwidth}
    \centering
    \begin{tabular}{crcl}
    \Xhline{3\arrayrulewidth}
    \# & Score & Range & System \\
    \hline
    1 & 0.440 & 1     & GPT-3.5 \\
    \hline
    2 & 0.304 & 2     & REF-F \\
    \hline
    3 & 0.186 & 3-5   & TransGEC \\
      & 0.169 & 3-5   & T5\\
      & 0.134 & 4-7   & BERT-fuse \\
      & 0.102 & 5-8   & Riken-Tohoku \\
      & 0.095 & 5-8  & REF-M \\
      & 0.054 & 7-9  & UEDIN-MS \\
      & 0.021 & 8-10 & PIE \\
      & -0.007 & 9-10 & GECToR-BERT \\
    \hline
    4 & -0.138 & 11-13 & LM-Critic \\
      & -0.145 & 11-13 & GECToR-ens \\
      & -0.179 & 11-14 & TemplateGEC \\
      & -0.227 & 13-14 & BART \\
    \hline
    5 & -0.809 & 15   & INPUT \\
    \Xhline{3\arrayrulewidth}
    \end{tabular}
    \subcaption{GPT-4-E + Difficulty}
    \hspace{1mm}
  \end{minipage}
  }
  \hspace{7mm}
  \scalebox{0.7}[0.7]{
  \begin{minipage}[b]{0.4\textwidth}
    \centering
    \begin{tabular}{crcl}
    \Xhline{3\arrayrulewidth}
    \# & Score & Range & System \\
    \hline
    1 & 0.429 & 1     & GPT-3.5 \\
    \hline
    2 & 0.237 & 2-4   & REF-F \\
      & 0.198 & 2-4   & TransGEC \\
      & 0.167 & 3-5   & T5 \\
      & 0.118 & 4-8   & REF-M \\
      & 0.107 & 4-8   & BERT-fuse \\
      & 0.093 & 5-9  & Riken-Tohoku \\
      & 0.075 & 6-10  & UEDIN-MS\\
      & 0.064 & 6-10 & GECToR-BERT \\
      & 0.026 & 8-10 & PIE \\
    \hline
    3 & -0.129 & 11-13 & LM-Critic \\
      & -0.130 & 11-13 & GECToR-ens \\
      & -0.163 & 11-13   & TemplateGEC \\
    \hline
    4 & -0.293 & 14     & BART \\
    \hline
    5 & -0.798 & 15   & INPUT \\
    \Xhline{3\arrayrulewidth}
    \end{tabular}
    \subcaption{GPT-4-E + Impact}
    \hspace{1mm}
  \end{minipage}
  }
  \hspace{7mm}
  \scalebox{0.7}[0.7]{
  \begin{minipage}[b]{0.4\textwidth}
    \centering
    \begin{tabular}{crcl}
    \Xhline{3\arrayrulewidth}
    \# & Score & Range & System \\
    \hline
    1 & 0.104 & 1-4   & PIE \\
      & 0.094 & 1-5   & REF-M \\
      & 0.084 & 1-7   & GPT-3.5 \\
      & 0.058 & 2-7   & BERT-fuse \\
      & 0.052 & 2-8   & GECToR-ens \\
      & 0.042 & 3-8   & TransGEC \\
      & 0.019 & 4-10  & UEDIN-MS \\
      & 0.010 & 5-11  & Riken-Tohoku \\
      & -0.017 & 7-11 & GECToR-BERT \\
      & -0.019 & 7-11 & T5 \\
      & -0.034 & 8-12 & INPUT \\
      & -0.087 & 10-15 & REF-F \\
      & -0.099 & 12-15 & BART \\
      & -0.102 & 12-15 & TemplateGEC \\
      & -0.104 & 12-15 & LM-Critic \\
    \Xhline{3\arrayrulewidth}
    \end{tabular}
    \subcaption{Llama 2-S}
    \hspace{1mm}
  \end{minipage}
  }
  \hspace{7mm}
  \scalebox{0.7}[0.7]{
  \begin{minipage}[b]{0.4\textwidth}
    \centering
    \begin{tabular}{crcl}
    \Xhline{3\arrayrulewidth}
    \# & Score & Range & System \\
    \hline
    1 & 0.236 & 1     & TransGEC \\
    \hline
    2 & 0.170 & 2-5   & T5 \\
      & 0.143 & 2-6   & UEDIN-MS \\
      & 0.141 & 2-6   & REF-M \\
      & 0.116 & 2-7   & GPT-3.5 \\
      & 0.095 & 4-7   & Riken-Tohoku \\
      & 0.048 & 6-9   & GECToR-BERT \\
      & 0.038 & 6-9  & BERT-fuse \\
      & -0.004 & 8-10 & PIE \\
      & -0.044 & 9-11 & GECToR-ens \\
      & -0.080 & 10-13 & REF-F \\
      & -0.093 & 10-13 & LM-Critic \\
      & -0.141 & 12-14   & BART \\
      & -0.165 & 13-14   & TemplateGEC \\
    \hline
    3 & -0.458 & 15   & INPUT \\
    \Xhline{3\arrayrulewidth}
    \end{tabular}
    \subcaption{GPT-3.5-S}
    \hspace{1mm}
  \end{minipage}
  }
  \hspace{7mm}
  \scalebox{0.7}[0.7]{
  \begin{minipage}[b]{0.4\textwidth}
    \centering
    \begin{tabular}{crcl}
    \Xhline{3\arrayrulewidth}
    \# & Score & Range & System \\
    \hline
    1 & 0.658 & 1     & GPT-3.5 \\
    \hline
    2 & 0.542 & 2     & REF-F \\
    \hline
    3 & 0.203 & 3-4   & TransGEC \\
      & 0.187 & 3-5   & T5 \\
      & 0.145 & 4-6   & BERT-fuse \\
      & 0.091 & 6-7   & Riken-Tohoku \\
      & 0.074 & 6-7  & REF-M \\
    \hline
    4 & 0.009 & 8-9  & UEDIN-MS \\
      & -0.032 & 8-10 & GECToR-BERT \\
      & -0.085 & 9-11 & PIE \\
      & -0.102 & 10-11 & LM-Critic \\
    \hline
    5 & -0.238 & 12-14 & TemplateGEC \\
      & -0.258 & 12-14 & GECToR-ens \\
      & -0.293 & 13-14  & BART \\
    \hline
    6 & -0.901 & 15   & INPUT \\
    \Xhline{3\arrayrulewidth}
    \end{tabular}
    \subcaption{GPT-4-S}
    \hspace{1mm}
  \end{minipage}
  }
  \hspace{7mm}
  \scalebox{0.7}[0.7]{
  \begin{minipage}[b]{0.4\textwidth}
    \centering
    \begin{tabular}{crcl}
    \Xhline{3\arrayrulewidth}
    \# & Score & Range & System \\
    \hline
    1 & 0.673 & 1-2   & GPT-3.5 \\
      & 0.636 & 1-2   & REF-F \\
    \hline
    2 & 0.194 & 3-4   & TransGEC \\
      & 0.184 & 3-4   & T5 \\
    \hline
    3 & 0.121 & 5-7   & BERT-fuse \\
      & 0.090 & 5-7   & Riken-Tohoku \\
      & 0.082 & 5-7  & REF-M \\
      & 0.022 & 7-8  & UEDIN-MS \\
    \hline  
    4 & -0.074 & 9-11 & LM-Critic \\
      & -0.076 & 9-11 & GECToR-BERT \\
      & -0.118 & 9-11 & PIE \\
    \hline
    5 & -0.213 & 12-13 & TemplateGEC \\
      & -0.238 & 12-13 & GECToR-ens \\
    \hline
    6 & -0.309 & 14   & BART \\
    \hline
    7 & -0.974 & 15   & INPUT \\
    \Xhline{3\arrayrulewidth}
    \end{tabular}
    \subcaption{GPT-4-S + Grammaticality}
    \hspace{1mm}
  \end{minipage}
  } 
  \hspace{7mm}
  \scalebox{0.7}[0.7]{
  \begin{minipage}[b]{0.4\textwidth}
    \centering
    \begin{tabular}{crcl}
    \Xhline{3\arrayrulewidth}
    \# & Score & Range & System \\
    \hline
    1 & 0.721 & 1     & GPT-3.5 \\
    \hline
    2 & 0.648 & 2     & REF-F \\
    \hline
    3 & 0.230 & 3-4   & TransGEC \\
      & 0.178 & 3-5   & T5 \\
      & 0.122 & 4-6   & BERT-fuse \\
      & 0.115 & 5-7   & REF-M \\
      & 0.063 & 6-7  & Riken-Tohoku \\
    \hline
    4 & -0.007 & 8-9  & UEDIN-MS \\
      & -0.058 & 8-11 & PIE \\
      & -0.066 & 9-11 & GECToR-BERT \\
      & -0.102 & 9-11 & LM-Critic \\
    \hline
    5 & -0.264 & 12-14 & GECToR-ens \\
      & -0.271 & 12-14   & TemplateGEC \\
      & -0.308 & 12-14   & BART \\
    \hline
    6 & -1.002 & 15   & INPUT \\
    \Xhline{3\arrayrulewidth}
    \end{tabular}
    \subcaption{GPT-4-S + Fluency}
    \hspace{1mm}
  \end{minipage}
  }
  \hspace{7mm}
  \scalebox{0.7}[0.7]{
  \begin{minipage}[b]{0.4\textwidth}
    \centering
    \begin{tabular}{crcl}
    \Xhline{3\arrayrulewidth}
    \# & Score & Range & System \\
    \hline
    1 & 0.653 & 1-2   & REF-F \\
      & 0.601 & 1-2   & GPT-3.5 \\
    \hline
    2 & 0.242 & 3-4   & T5 \\
      & 0.209 & 3-4   & TransGEC \\
    \hline
    3 & 0.135 & 5-6   & REF-M \\
      & 0.106 & 5-7   & BERT-fuse \\
      & 0.071 & 6-7  & Riken-Tohoku \\
    \hline
    4 & 0.011 & 8  & UEDIN-MS \\  
    \hline
    5 & -0.067 & 9-10 & GECToR-BERT \\
      & -0.106 & 9-11 & LM-Critic \\
      & -0.123 & 10-11 & PIE \\
    \hline
      & -0.225 & 12-13 & TemplateGEC \\
      & -0.255 & 12-13 & GECToR-ens \\
    \hline
    7 & -0.317 & 14     & BART \\
    \hline
    8 & -0.935 & 15   & INPUT \\
    \Xhline{3\arrayrulewidth}
    \end{tabular}
    \subcaption{GPT-4-S + Meaning Preservation}
  \end{minipage}
  }
  \caption{LLM rankings generated using Trueskill based on pairwise judgments made by LLMs. GPT-4 ranks fluent corrections (REF-F, GPT-3.5) highly, resulting in these rankings that closely resemble human ranking.} 
  \label{tab:llm_rankings}
\end{table*}

\section{Tendency of LLM scoring}\label{sec:llm_scoring}
\begin{figure*}[t]
\centering
\includegraphics[width=16cm]{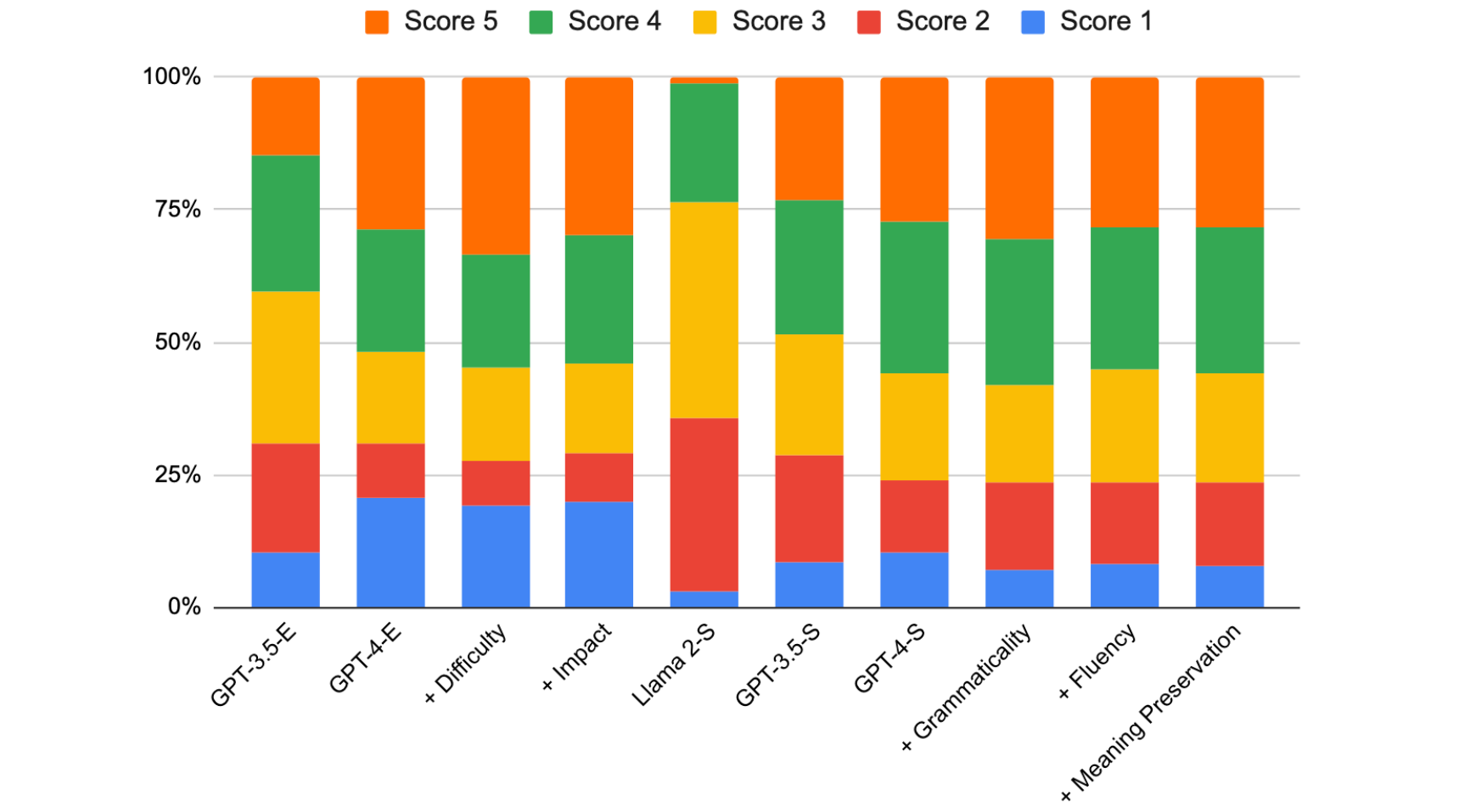} 
\caption{
The distribution of scores assigned by LLMs on a 5-point scale.
It can be observed that as the LLM scale increases, there is a tendency to assign higher scores (4 or 5 points).
Based on our meta-evaluation results indicating better correlation with human judgments as the scale increases, it is suggested that smaller LLMs may underestimate corrections judged to be good by humans.
}
\label{fig:llm_scoring}
\end{figure*}

The distribution of scores assigned by LLMs to corrected sentences is shown in Figure~\ref{fig:llm_scoring}.
As the LLM scale increases, there is a tendency to assign higher scores (4 or 5 points).
Based on our meta-evaluation results, which suggest that higher LLM scales are associated with higher correlations with human evaluations, smaller LLMs may underestimate corrections judged to be good by humans.
Llama 2-S tends to avoid extreme scores such as 1 or 5 points and shows a high degree of score overlap, making it difficult to compare more detailed corrected sentences.

\end{document}